\documentclass[letterpaper,10pt,conference]{ieeeconf}

\IEEEoverridecommandlockouts

\usepackage{cite}

\usepackage{algorithm}
\usepackage{algorithmic}

\usepackage{amsmath}
\usepackage{amsfonts}
\usepackage{amssymb}

\usepackage{graphicx}
\usepackage{subcaption}
\usepackage{float}
\usepackage{tikz}

\usepackage{booktabs}
\usepackage{multirow}
\usepackage{makecell}
\usepackage{siunitx}

\usepackage[utf8]{inputenc}
\usepackage{textcomp}
\usepackage{xcolor}

\usepackage[
    hidelinks,
    bookmarks=false
]{hyperref}

\def\BibTeX{{\rm B\kern-.05em{\sc i\kern-.025em b}\kern-.08em
T\kern-.1667em\lower.7ex\hbox{E}\kern-.125emX}}

\title{\LARGE \bf
Can Knowledge Transfer Parameters Be Learned?\\
LePoKet for Efficient Robotic Vision
}

\author{
Yanick C. Tchenko$^{1}$,
Felix Mohr$^{2}$,
Hicham H. Abdelkader$^{1}$,
and Hedi Tabia$^{1}$%
\thanks{$^{1}$Université Paris-Saclay, IBISC Laboratory, France.}%
\thanks{$^{2}$Universidad de La Sabana, Colombia.}%
}

\begin{document}

\maketitle
\thispagestyle{empty}
\pagestyle{empty}


\begin{abstract}
Efficient perception is central to robotic systems operating under constrained computation, memory, and latency budgets. Knowledge transfer from larger pretrained models offers a practical route to stronger compact perception networks, but existing approaches commonly rely on fixed distillation objectives or manually designed interaction mechanisms. Building on Hereditary Knowledge Transfer (HKT), we propose LePoKet (Learnable Parameter Optimization for Knowledge Transfer), a structural transfer framework that embeds knowledge inheritance directly into the forward computation. LePoKet introduces a block-wise Extract--Transform--Mix interface whose interaction parameters are optimized jointly with the child network through a Learnable Genetic Attention (LGA) operator, without auxiliary distillation losses or temperature scaling. We first characterize the mechanism on CIFAR-10 and CIFAR-100 with ResNet parent--child pairs, obtaining relative error reductions of 24.57\% and 25.1\% over standard child training. We then evaluate LePoKet in dense motion estimation by integrating it into a compact RAFT-based optical-flow model trained only on FlyingChairs and FlyingThings3D. LePoKet improves the compact RAFT baseline from 2.21 to 1.92 EPE on Sintel Clean, from 3.35 to 3.01 on Sintel Final, and from 7.51 to 6.39 on KITTI. A direct comparison with HKT further shows that LePoKet improves CIFAR-10 accuracy from 92.40\% to 93.40\% and, on C+T optical flow, trades a negligible Sintel-Clean difference (1.90/1.91 versus 1.92) for lower Sintel-Final and KITTI error. These results indicate that learnable structural transfer is applicable to both recognition and motion perception and motivates its use for efficient robotic vision.
\textbf{Code:} \url{https://github.com/christian-tchenko/LePoKet}

\end{abstract}


\section{Introduction}

Modern robotic systems increasingly rely on deep perception models for
motion estimation, scene understanding, navigation, and interaction with
dynamic environments. At the same time, deployment on mobile robots,
autonomous platforms, and embedded devices imposes stringent constraints
on computation, memory, and latency. This motivates methods that can exploit
knowledge acquired by large pretrained models while producing compact
networks suitable for resource-constrained perception.

Knowledge Distillation (KD) \cite{hinton2015distilling} is a widely used
framework for transferring information from a large teacher to a compact
student. Classical KD transfers softened output distributions through an
auxiliary objective, while subsequent methods extend distillation to
intermediate representations, including FitNets
\cite{romero2014fitnets}, Attention Transfer (AT)
\cite{zagoruyko2017attention}, Similarity-Preserving (SP) transfer
\cite{tung2019similarity}, and Contrastive Representation Distillation
(CRD) \cite{tian2019crd}. More recent approaches refine the transfer
objective through decoupled supervision \cite{zhao2022dkd},
review-based strategies \cite{chen2021reviewkd}, or feature-based
alignment \cite{cho2019overhaul}. Despite their effectiveness, these
approaches remain predominantly objective-driven: knowledge transfer is
induced by additional terms that encourage the student to reproduce
selected teacher outputs, features, or relations.

Hereditary Knowledge Transfer (HKT)
\cite{tchenko2026hereditary} introduced a different perspective in which
knowledge inheritance is incorporated structurally into the interaction
between a parent and a child network. HKT organizes this interaction through
an Extract--Transform--Mix (ETM) mechanism and uses Genetic Attention to
selectively combine inherited and child representations. This structural
view demonstrated that knowledge transfer need not be formulated solely as
teacher--student imitation and provides the direct methodological foundation
for the present work.

Building on this principle, we investigate a more specific question:
\emph{can the parameters governing the transfer operation themselves be
learned jointly with the child network?} We introduce \textbf{LePoKet}
(\emph{Learnable Parameter Optimization for Knowledge Transfer}), which
retains the structural ETM formulation while replacing the prescribed
interaction with a differentiable, parameterized transfer operator.
Its Learnable Genetic Attention (LGA) mechanism learns projections,
compatibility scores, gating parameters, and transfer intensity directly
from the task objective. The pretrained parent remains frozen, whereas
the child network and the parameters of the transfer interface are jointly
optimized.

This distinction is important. LePoKet does not introduce another
teacher--student discrepancy objective. Instead, parent information enters
the child's forward computation, and the usefulness of that information is
determined indirectly by its contribution to the task loss. Consequently,
the optimization can learn not only the child representation but also
\emph{how} inherited information should modify that representation.

We evaluate this formulation in two complementary settings. First,
CIFAR-10 and CIFAR-100 classification provide controlled recognition
benchmarks in which the effect of learnable transfer can be isolated using
ResNet parent--child pairs. Second, we integrate LePoKet into a compressed
RAFT optical-flow estimator \cite{teed2020raft}. Optical flow provides a
dense motion-perception setting relevant to robotic and autonomous systems
and allows us to examine whether the same transfer principle extends from
image-level recognition to structured pixel-level prediction.

The main contributions of this work are:
\begin{itemize}
    \item We extend hereditary structural transfer by formulating the
    parameters of the parent--child interaction as jointly learnable
    components of the network.
    
    \item We introduce Learnable Genetic Attention (LGA), which combines
    projected parent and child representations through learned similarity,
    nonlinear gating, and residual transfer intensity.
    
    \item We experimentally separate the benefit of structural inheritance
    from that of learning the transfer interface through direct
    HKT--LePoKet comparisons on image classification and dense optical flow.
\end{itemize}

Across the reported experiments, LePoKet consistently improves the compact
student baseline. The comparison with HKT further shows that learning the
transfer interface changes the behavior of structural inheritance across
tasks and evaluation conditions, motivating parameterized hereditary
transfer as a flexible extension of the original HKT formulation.

\section{Related Work}

\subsection{Knowledge Distillation}

Knowledge Distillation (KD) \cite{hinton2015distilling} transfers softened
teacher predictions to a student through temperature-scaled supervision.
FitNets \cite{romero2014fitnets} extend this principle to intermediate
representations, while Attention Transfer (AT)
\cite{zagoruyko2017attention} aligns attention maps.
Similarity-Preserving (SP) distillation \cite{tung2019similarity} and
Contrastive Representation Distillation (CRD) \cite{tian2019crd} further
transfer relational information in feature space. More recent approaches
include Decoupled Knowledge Distillation (DKD) \cite{zhao2022dkd},
review-based feature transfer \cite{chen2021reviewkd}, and feature-oriented
alignment strategies \cite{cho2019overhaul}.

Although these methods differ in the information transferred, they
primarily express teacher--student interaction through additional training
objectives. LePoKet instead places the transfer mechanism inside the
forward computation and optimizes its parameters through the task loss.

\subsection{Knowledge Transfer for Dense Motion Perception}

Knowledge transfer has also been investigated for dense prediction.
DRAFT \cite{tchenko2023draft}, for example, applies distillation to
RAFT-based optical-flow estimation using teacher supervision to improve
compact flow models. Optical flow is particularly relevant to robotic and
autonomous perception because it provides dense information about image
motion and scene dynamics.

Our RAFT experiment uses this setting to test a different transfer
mechanism. Rather than introducing an additional distillation objective,
LePoKet injects aligned parent information through its learnable transfer
interface while preserving the standard task supervision.

\subsection{Hereditary Knowledge Transfer}

HKT \cite{tchenko2026hereditary} introduced hereditary knowledge transfer
as a modular mechanism for transferring representations between neural
networks. Its Extract--Transform--Mix formulation separates three
operations: selecting transferable parent information, adapting it to the
child representation, and selectively integrating it into the child
network. Genetic Attention regulates this inheritance process.

This structural formulation is the direct precursor of LePoKet. We retain
the ETM principle but parameterize the interaction so that the transfer
operation can itself be optimized jointly with the child. In particular,
LePoKet introduces learnable projections, a nonlinear compatibility gate,
and a learnable transfer coefficient within LGA. The resulting formulation
therefore preserves hereditary transfer while shifting part of the design
choice from a prescribed mechanism to task-driven parameter learning.

This relationship also motivates our experimental comparison. Rather than
treating HKT as an unrelated baseline, we use it to distinguish two effects:
the gain obtained by introducing a structural inheritance path and the
additional effect obtained when the parameters governing that path are
learned.

\subsection{Attention-Based Transfer}

Attention mechanisms \cite{vaswani2017attention} provide a general means
of weighting interactions between representations and have also been used
for knowledge distillation \cite{zagoruyko2017attention}. In many such
methods, attention defines the representation or relation that an auxiliary
loss attempts to align.

LePoKet uses attention differently. LGA is part of the parent--child
forward interaction itself. It combines a projected similarity term with
a learnable nonlinear compatibility function and uses the resulting
coefficient to regulate a residual update of the child representation.
The attention mechanism therefore determines the transfer operation rather
than defining an additional imitation target.

\subsection{Positioning of LePoKet}

Table~\ref{tab:positioning} summarizes the distinction between conventional
distillation, hereditary structural transfer, and LePoKet. The central
difference is not whether teacher information is adaptive, but
\emph{where the transfer mechanism is represented and optimized}.
Conventional distillation primarily encodes transfer in the objective;
HKT introduces an explicit structural inheritance path; LePoKet extends
that path with parameters that are jointly optimized with the child through
the task objective.

\begin{table*}[t]
\centering
\caption{Positioning of LePoKet relative to representative knowledge-transfer
paradigms. The distinction concerns how teacher/parent information enters
training and whether the parameters of the structural transfer operator are
jointly learned with the child.}
\label{tab:positioning}
\begin{tabular}{lcccc}
\toprule
Method family &
Transfer source &
Forward inheritance &
Transfer-specific supervision &
Learnable transfer operator \\
\midrule
Logit KD \cite{hinton2015distilling}
& Outputs & No & Yes & -- \\

Feature/attention KD
\cite{romero2014fitnets,zagoruyko2017attention}
& Features & No & Yes & -- \\

Relational KD
\cite{tung2019similarity,tian2019crd}
& Relations & No & Yes & -- \\

HKT \cite{tchenko2026hereditary}
& Representations & Yes & Yes & Prescribed/structured \\

\textbf{LePoKet (ours)}
& Representations & \textbf{Yes} & \textbf{No} & \textbf{Yes} \\
\bottomrule
\end{tabular}
\end{table*}


\section{LePoKet Method}\label{sec:method}

\subsection{Problem Formulation}
We consider a frozen pretrained \emph{parent} network $P$ and a compact \emph{child} network $C$. Both are decomposed into $L$ functional stages,
\begin{equation}
P=\{p_\ell\}_{\ell=1}^{L}, \qquad C=\{c_\ell\}_{\ell=1}^{L},
\end{equation}
with intermediate representations $\mathbf z_\ell^P$ and $\mathbf z_\ell^C$. The parent is used only as a source of representations during training; its parameters $\theta_P$ remain fixed. The child parameters $\theta_C$ are optimized for the downstream task.

Conventional distillation typically introduces an auxiliary discrepancy between parent and child outputs or features. LePoKet instead asks the transfer interface to learn how parent information should enter the child computation. The task objective therefore remains the only supervision used to optimize the child and the transfer parameters.

\begin{figure*}[t]
    \centering
    \includegraphics[width=0.78\linewidth]{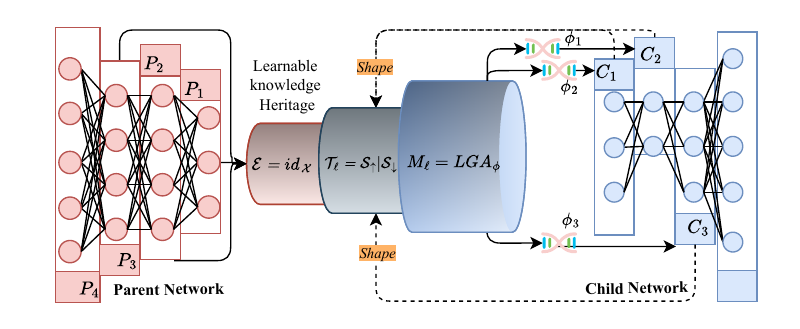}
    \caption{Conceptual overview of LePoKet. A frozen parent transfers intermediate representations to a child through a block-wise learnable Extract--Transform--Mix (ETM) interface. Extraction selects a parent representation, transformation provides structural alignment when required, and mixing is performed by Learnable Genetic Attention (LGA). Only the child and transfer parameters are optimized by the task loss.}
    \label{fig:concept}
\end{figure*}

\subsection{From Structural Inheritance to Learnable Transfer}\label{sec:hkt_to_lepoket}
The starting point is HKT \cite{tchenko2026hereditary}, which injects parent information through an Extract--Transform--Mix interface and uses Genetic Attention to select inherited information. HKT establishes that a structural inheritance path can improve a compact child, but its ETM interaction is not jointly parameterized and optimized as the task-driven interface proposed here; its original training also aggregates core and inheritance supervision. LePoKet retains the ETM abstraction while replacing the prescribed interaction with learnable projections, a learnable gate, and a learnable transfer intensity, all optimized through the standard task objective. This HKT$\rightarrow$LePoKet transition isolates the benefit of learning the transfer path itself.

LePoKet generalizes this mechanism by parameterizing the transfer operation. At stage $\ell$,
\begin{equation}
\tilde{\mathbf z}_\ell^C = \mathcal M_\ell\!\left(\mathbf z_\ell^C,
\mathcal T_\ell(\mathcal E_\ell(\mathbf z_\ell^P));\phi_\ell\right),
\label{eq:etm}
\end{equation}
where $\mathcal E_\ell$ extracts a parent representation, $\mathcal T_\ell$ aligns it to the child representation, $\mathcal M_\ell$ performs the mixing, and $\phi_\ell$ contains learnable transfer parameters. In the ResNet experiment, extraction is the identity,
\begin{equation}
\mathcal E_\ell(\mathbf z_\ell^P)=\mathbf z_\ell^P,
\end{equation}
and the pooled representations already share the required dimensionality. For heterogeneous stages, $\mathcal T_\ell$ can perform channel or spatial alignment before mixing.

This formulation separates three questions that are otherwise entangled in a distillation loss: \emph{what} parent representation is used, \emph{how} it is made compatible with the child, and \emph{how much} of it is inherited for a particular input. LePoKet focuses learnability on the last two operations while keeping the parent fixed.

\subsection{Learnable Genetic Attention}\label{sec:lga}
LePoKet instantiates $\mathcal M_\ell$ with Learnable Genetic Attention (LGA). Let $\mathbf q,\mathbf k,\mathbf v\in\mathbb R^d$ denote the child query and parent-derived key/value representations after alignment. Learnable projections are
\begin{equation}
\mathbf Q=W_q\mathbf q,\qquad \mathbf K=W_k\mathbf k,\qquad \mathbf V=W_v\mathbf v,
\end{equation}
where $W_q,W_k,W_v$ are trainable linear mappings. LGA combines a global similarity term
\begin{equation}
a_1=\sigma\!\left(\frac{\langle\mathbf Q,\mathbf K\rangle}{\sqrt d}\right)
\end{equation}
with a nonlinear compatibility gate
\begin{equation}
a_2=\sigma\!\left(\mathrm{MLP}(\mathbf Q\odot\mathbf K)\right),
\end{equation}
where $\odot$ is element-wise multiplication and $\sigma$ is the sigmoid function. The inheritance coefficient is
\begin{equation}
a=\frac{1}{2}(a_1+a_2).
\label{eq:attention}
\end{equation}
The first term captures global alignment through a scaled dot product, whereas the second can model nonlinear feature-wise compatibility. Their average provides a bounded, data-dependent coefficient without introducing a separate attention supervision signal.

The child representation is updated through a residual interpolation toward the transformed parent value:
\begin{equation}
\tilde{\mathbf z}_\ell^C=\mathbf q+\lambda_\ell a\,(\mathbf V-\mathbf q),
\label{eq:fusion}
\end{equation}
where $\lambda_\ell$ is learnable. Equation~\eqref{eq:fusion} makes the role of the transfer explicit. The vector $(\mathbf V-\mathbf q)$ gives a parent-directed correction, while $\lambda_\ell a$ determines its magnitude. Small compatibility leaves the child close to its own representation; larger compatibility allows stronger inheritance.

\subsection{Optimization and Gradient Flow}
Let $\phi=\{W_q,W_k,W_v,\lambda,\text{MLP weights}\}$ denote all LePoKet parameters. For a sample $(x,y)$, the parent forward pass produces frozen source features and the child forward pass receives the fused representations. Training solves
\begin{equation}
\min_{\theta_C,\phi}\;\mathbb E_{(x,y)\sim\mathcal D}\left[
\mathcal L_{\mathrm{task}}(C_{\mathrm{LePoKet}}(x;\theta_C,\phi),y)\right].
\label{eq:objective}
\end{equation}
Gradients from $\mathcal L_{\mathrm{task}}$ propagate through Eq.~\eqref{eq:fusion}, the LGA projections and gate, and the child network, but not into $\theta_P$. Consequently, no temperature, teacher-logit loss, feature-matching loss, or auxiliary loss weight is introduced. The transfer interface is selected indirectly according to whether its inherited representation improves the downstream task objective.

\begin{algorithm}[t]
\caption{LePoKet Training}
\label{alg:lepoket}
\begin{algorithmic}[1]
\STATE Initialize $\theta_C,\phi$; freeze $\theta_P$
\FOR{each minibatch $(x,y)$}
    \STATE $\{\mathbf z_\ell^P\}_{\ell=1}^{L}\leftarrow P(x)$
    \STATE compute child representations $\{\mathbf z_\ell^C\}_{\ell=1}^{L}$
    \FOR{$\ell=1$ to $L$}
        \STATE $\boldsymbol\tau_\ell\leftarrow\mathcal T_\ell(\mathcal E_\ell(\mathbf z_\ell^P))$
        \STATE $a_\ell\leftarrow g(\mathbf z_\ell^C,\boldsymbol\tau_\ell;\phi_\ell)$
        \STATE $\mathbf z_\ell^C\leftarrow\mathbf z_\ell^C+\lambda_\ell a_\ell(\mathbf V_\ell-\mathbf z_\ell^C)$
    \ENDFOR
    \STATE $\hat y\leftarrow C_{\mathrm{head}}(\mathbf z_L^C)$
    \STATE $\mathcal L\leftarrow\mathcal L_{\mathrm{task}}(\hat y,y)$
    \STATE update $\theta_C,\phi$ using $\nabla\mathcal L$
\ENDFOR
\end{algorithmic}
\end{algorithm}

\subsection{Architecture-Agnostic Instantiation}
The ETM decomposition is intended to decouple the transfer rule from a particular backbone. If parent and child features have equal dimensions, $\mathcal T_\ell$ may be the identity. If channel dimensions differ, a projection can map the parent representation to the child space; if spatial resolutions differ, resizing can be included before projection. The same LGA mixing rule can then operate on the aligned representations.

In the ResNet setting used below, transfer is applied to pooled representations and the dimensions already match. In the RAFT setting, parent and compact child representations require explicit alignment before the LGA interaction. These two cases test the same formulation under both homogeneous and heterogeneous parent--child architectures.

\subsection{Design Properties for Robotic Perception}
Three properties motivate the use of LePoKet for compact perception. First, the parent is frozen, so the source model does not need to be co-adapted with the child. Second, transfer is data-dependent: the inheritance coefficient is computed from the current parent--child representations rather than fixed globally. Third, the optimization remains task-driven. This is useful when the downstream objective already has a well-established form, as in classification cross-entropy or optical-flow endpoint error, because the transfer mechanism can be introduced without redesigning that objective.

These properties do not by themselves establish deployment efficiency. The present study evaluates predictive behavior of the compact models; latency, memory, energy, and hardware-specific measurements are outside the current experiments and are identified explicitly as future deployment evaluation.

\subsection{Scope and Expected Failure Modes}
LePoKet assumes that a useful correspondence can be established between at least one parent representation and a child representation. If the selected stages encode incompatible information, an alignment transform can make their tensor shapes compatible but cannot guarantee semantic compatibility. The learned gate can reduce inheritance when representations disagree, yet it cannot create useful parent information that is absent from the selected feature. Stage selection therefore remains an architectural design choice.

A second consideration is the quality of the parent. Because the parent is frozen, systematic errors in its representation are not corrected through parent updates. The residual form in Eq.~\eqref{eq:fusion} is intended to preserve a direct child path, so the child is not forced to replace its representation with the parent value. Nevertheless, evaluating robustness to weak or mismatched parents is outside the present experiments.

Finally, LePoKet adds parameters and parent-side computation during transfer training. The current contribution concerns whether these transfer parameters can be optimized effectively, not a claim of zero-cost training. For deployment-oriented robotics, the relevant follow-up is to separate training-time transfer cost from the cost of the compact model used at inference and to measure both on target hardware.

\subsection{Interpretation of the Transfer Dynamics}\label{sec:dynamics}
Equation~\eqref{eq:fusion} also clarifies how LePoKet differs from simply copying a parent feature. Define $\alpha_\ell=\lambda_\ell a_\ell$. The update can be rewritten as
\begin{equation}
\tilde{\mathbf z}_\ell^C=(1-\alpha_\ell)\mathbf q+\alpha_\ell\mathbf V.
\label{eq:interp}
\end{equation}
Thus, when $\alpha_\ell$ lies between zero and one, the fused representation is an interpolation between the child's current state and the projected parent state. When the learned interaction approaches zero, Eq.~\eqref{eq:interp} reduces to the ordinary child path. This gives the model a mechanism for suppressing transfer on inputs for which the parent and child representations are judged incompatible. Conversely, a larger interaction increases the contribution of the parent-derived representation.

The distinction between $a_\ell$ and $\lambda_\ell$ is useful. The attention term is input-dependent because it is computed from $\mathbf Q$ and $\mathbf K$, whereas $\lambda_\ell$ is a learned transfer-intensity parameter associated with the interface. The product therefore combines a learned stage-level tendency to inherit with sample-dependent compatibility. In this sense, LePoKet does not assume that a fixed amount of parent information is equally useful for every example.

The residual form also preserves a direct optimization path for the child. The task loss can improve the child representation itself, the parent projection, or the gate that determines their interaction. This differs from a feature-matching objective that explicitly penalizes distance to a teacher representation: LePoKet is not optimized to make $\mathbf q$ equal to $\mathbf V$. Parent information is useful only insofar as the resulting fused representation reduces the downstream task loss.

This interpretation motivates the ablation in Sec.~\ref{sec:ablation}. Comparing the task-only child with fixed structural HKT tests whether an inheritance path is useful at all; comparing HKT with LePoKet then tests whether allowing the interaction to adapt provides additional benefit. The two comparisons answer different questions and should not be conflated. The magnitude of the HKT-to-LePoKet difference then indicates how much additional benefit is obtained by optimizing the transfer interface beyond introducing the structural inheritance path itself.

\section{Experiments}\label{sec:experiments}
We evaluate LePoKet on image classification and dense optical-flow estimation. Classification provides a controlled setting for isolating the effect of learnable inheritance, while optical flow tests the same mechanism on pixel-level motion perception relevant to robotic and autonomous systems. In all experiments the parent is frozen and no auxiliary distillation loss or temperature scaling is used.

\subsection{Evaluation Questions}
The experiments address three questions. \textbf{Q1:} Does a learnable transfer interface improve a compact child over standard task-only training? \textbf{Q2:} Does the mechanism extend from classification to dense motion estimation without changing its basic formulation? \textbf{Q3:} How much of the observed gain is attributable to structural inheritance itself, and how much to making the interaction parameters learnable? Q3 is examined by comparing the baseline child, structural HKT, and LePoKet.

\subsection{Experimental Setup}
\textbf{Image classification.} We use CIFAR-10 and CIFAR-100 \cite{krizhevsky2009learning}, each containing 50,000 training and 10,000 test images of size $32\times32$. ResNet110 \cite{he2016resnet} is the frozen parent and ResNet20 the child. The parent reaches 93.6\% Top-1 accuracy on CIFAR-10. The child is trained with SGD, momentum 0.9, weight decay $10^{-4}$, initial learning rate 0.01 with scheduled decay, and batch size 128. Augmentation uses random cropping with padding and horizontal flipping.

\begin{figure*}[t]
    \centering
    \includegraphics[width=0.80\linewidth]{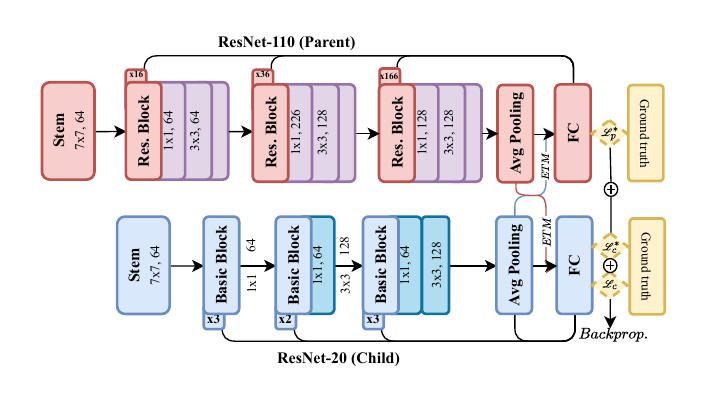}
    \caption{LePoKet in the ResNet experiment. ResNet110 is the frozen parent and ResNet20 is the child. Transfer is applied to the pooled representation through the learnable ETM/LGA interface.}
    \label{fig:resnethkt}
\end{figure*}

As illustrated in Fig.~\ref{fig:resnethkt}, transfer is applied at the final pooled representation. The projection dimension is 64, matching the feature dimension. The learnable set $\phi=\{W_q,W_k,W_v,\lambda,\mathrm{MLP}\}$ is optimized jointly with the child.

\textbf{Optical flow.} We integrate LePoKet into a compressed RAFT architecture. Training uses FlyingChairs \cite{Dosovitskiy2015FlyingChairs} and FlyingThings3D \cite{Mayer2016FlyingThings3D} (C+T) only, with no Sintel or KITTI fine-tuning. Evaluation is reported on Sintel Clean (S-C), Sintel Final (S-F), and KITTI endpoint error (K-E). This setting tests whether the transfer mechanism carries from global recognition features to dense motion representations.

\subsection{Metrics and Comparison Protocol}
For classification we report Top-1 accuracy and error
$E=100-\text{Top-1}$. Relative error reduction (RER) with
respect to the child baseline is
\begin{equation}
\mathrm{RER}=
\frac{E_{\mathrm{baseline}}-E_{\mathrm{model}}}
     {E_{\mathrm{baseline}}}\times100\%.
\end{equation}
For optical flow, lower endpoint error is better.

Literature methods in the classification tables are included as
contextual reference points; because their reported training protocols
can differ, they are not treated as controlled reproductions.
For HALOC~\cite{haloc2023}, the original work reports a
$0.07$ percentage-point accuracy improvement over an uncompressed
ResNet20 on CIFAR-10. For consistent presentation in
Table~\ref{tab:cifar10_comparison}, we therefore express this reported
improvement relative to our 91.25\% ResNet20 reference, yielding
91.32\%. The corresponding error and RER values are recomputed using
the definition above.

The controlled evidence in this paper is the comparison with the
corresponding compact baseline and the published HKT predecessor
\cite{tchenko2026hereditary}. For HKT, RER values in our tables are
likewise recomputed from the reported accuracies using the definition
above so that the comparison is internally consistent.

\subsection{CIFAR-10 Results}
\begin{table}[t]
\centering
\caption{CIFAR-10: ResNet110 $\rightarrow$ ResNet20.}
\label{tab:cifar10_comparison}
\begin{tabular}{lccc}
\toprule
Method & Top-1 (\%) & Error (\%) & RER (\%)\\
\midrule
Parent (R-110) & 93.6 & 6.4 & --\\
Baseline (R-20) & 91.25 & 8.75 & --\\
\midrule
HALOC \cite{haloc2023} & 91.32 & 8.68 & 0.80\\
SparseKD \cite{sparsekd2024} & 91.5 & 8.5 & 2.86\\
Low-Rank \cite{idelbayev2020lowrank} & 91.0 & 9.0 & -2.86\\
RegPrune \cite{vu2020regularization} & 91.2 & 8.8 & -0.57\\
\midrule
HKT \cite{tchenko2026hereditary} & 92.40 & 7.60 & 13.14\\
\textbf{LePoKet (Ours)} & \textbf{93.40} & \textbf{6.60} & \textbf{24.57}\\
\bottomrule
\end{tabular}
\end{table}

Table~\ref{tab:cifar10_comparison} shows that the ResNet20 baseline obtains 91.25\% accuracy. LePoKet reaches 93.40\%, an absolute gain of 2.15 percentage points and a relative error reduction of 24.57\%. The result answers Q1 positively in the controlled CIFAR-10 setting: learning the transfer interface improves the same compact child while retaining the standard task objective. Relative to the published HKT result (92.40\%), LePoKet adds 1.00 percentage point, indicating a substantial additional benefit from optimizing the transfer interface on this dataset.

The contextual methods in Table~\ref{tab:cifar10_comparison} span compression, sparsity, and knowledge-transfer approaches. We report their values to situate the operating range rather than to claim a strictly identical training protocol. This distinction is important because LePoKet's central comparison is with the task-only child and the structural HKT variant in Sec.~\ref{sec:ablation}.

\subsection{CIFAR-100 Results}
\begin{table}[t]
\centering
\caption{CIFAR-100: ResNet110 $\rightarrow$ ResNet20.}
\label{tab:cifar100_distill}
\begin{tabular}{lccc}
\toprule
Method & Top-1 (\%) & Error (\%) & RER (\%)\\
\midrule
Baseline (R-20) & 65.3 & 34.7 & --\\
KD \cite{hinton2015distilling} & 66.0 & 34.0 & 2.0\\
AT \cite{zagoruyko2017attention} & 66.2 & 33.8 & 2.6\\
SP \cite{tung2019similarity} & 66.3 & 33.7 & 2.9\\
NORM \cite{cho2019overhaul} & 66.6 & 33.4 & 3.7\\
CRD \cite{tian2019crd} & 67.0 & 33.0 & 4.9\\
DKD \cite{zhao2022dkd} & 67.4 & 32.6 & 6.1\\
\midrule
\textbf{LePoKet (Ours)} & \textbf{74.01} & \textbf{25.99} & \textbf{25.1}\\
\bottomrule
\end{tabular}
\end{table}

On CIFAR-100, Table~\ref{tab:cifar100_distill} reports a larger separation from the compact baseline. ResNet20 obtains 65.3\%, whereas LePoKet reaches 74.01\%. The corresponding error decreases from 34.7\% to 25.99\%, giving a 25.1\% RER. This result indicates that the same transfer formulation remains effective when the recognition problem contains more classes and the child baseline has substantially larger error.

As in the CIFAR-10 table, the published distillation values are contextual rather than a claim of identical reproduction. The main observation is therefore not that LePoKet universally dominates every distillation implementation, but that task-driven optimization of the transfer interface yields a substantial improvement over the compact child under the reported setting.

\subsection{Dense Motion Perception with RAFT}
The optical-flow experiment addresses Q2. LePoKet is inserted at the feature interaction stage of a compressed RAFT model. Parent representations are extracted and aligned to the compact feature space before LGA mixing. The recurrent GRU-based refinement and cost-volume construction of RAFT are otherwise preserved. Transfer parameters and child parameters are optimized using the standard endpoint-error objective.

\begin{table}[t]
\centering
\caption{Optical-flow results with FlyingChairs + FlyingThings3D (C+T) training only. Lower is better.}
\label{tab:flow_ct}
\begin{tabular}{lccc}
\toprule
Method & S-C $\downarrow$ & S-F $\downarrow$ & K-E $\downarrow$\\
\midrule
RAFT ($\mathcal P$) \cite{teed2020raft} & 1.43 & 2.71 & 5.04\\
RAFT ($\mathcal C$) Baseline & 2.21 & 3.35 & 7.51\\
DRAFT \cite{tchenko2023draft} & 1.99 & 3.17 & 6.94\\
2HKT-RAFT \cite{tchenko2026hereditary} & 1.91 & 3.03 & 7.37\\
3HKT-RAFT \cite{tchenko2026hereditary} & 1.90 & 3.08 & 6.45\\
\midrule
\textbf{LePoKet (Ours)} & \textbf{1.92} & \textbf{3.01} & \textbf{6.39}\\
\bottomrule
\end{tabular}
\end{table}

Table~\ref{tab:flow_ct} shows consistent improvement over the compact RAFT baseline: 2.21$\rightarrow$1.92 on Sintel Clean, 3.35$\rightarrow$3.01 on Sintel Final, and 7.51$\rightarrow$6.39 on KITTI. The model is trained only on C+T, so these results do not rely on benchmark-specific fine-tuning. Relative to DRAFT under the reported protocol, LePoKet also gives lower error on all three metrics. The HKT comparison is more informative: LePoKet is slightly worse on Sintel Clean (1.92 versus 1.91/1.90), but improves Sintel Final (3.01 versus 3.03/3.08) and KITTI EPE (6.39 versus 7.37/6.45) relative to the reported 2HKT/3HKT variants. Thus, learning the interface changes the trade-off across dense-motion benchmarks rather than uniformly improving every metric. The experiment demonstrates that the same learnable structural-transfer principle can be applied beyond classification to dense motion estimation.

\subsection{Structural vs. Learnable Transfer}\label{sec:ablation}
To address Q3, we distinguish three configurations. \textbf{Baseline} is the compact model trained with its task loss. \textbf{HKT} is the published structural hereditary-transfer predecessor \cite{tchenko2026hereditary}, based on ETM and Genetic Attention with aggregated inheritance supervision. \textbf{LePoKet} retains the structural ETM principle but explicitly
parameterizes the transfer operator and jointly optimizes its
parameters through LGA using the task objective alone.

\begin{figure}[t]
    \centering
    \begin{minipage}{0.48\linewidth}
        \centering
        \includegraphics[width=\linewidth]{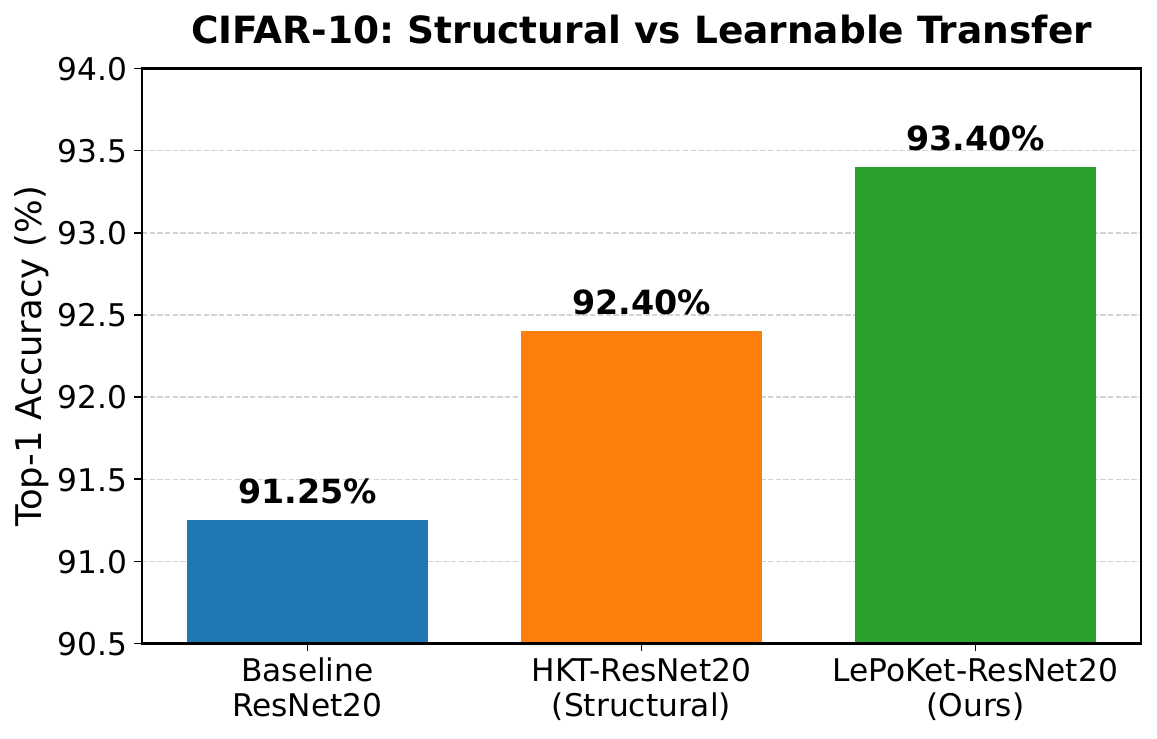}
        {\small (a) CIFAR-10 classification}
    \end{minipage}\hfill
    \begin{minipage}{0.48\linewidth}
        \centering
        \includegraphics[width=\linewidth]{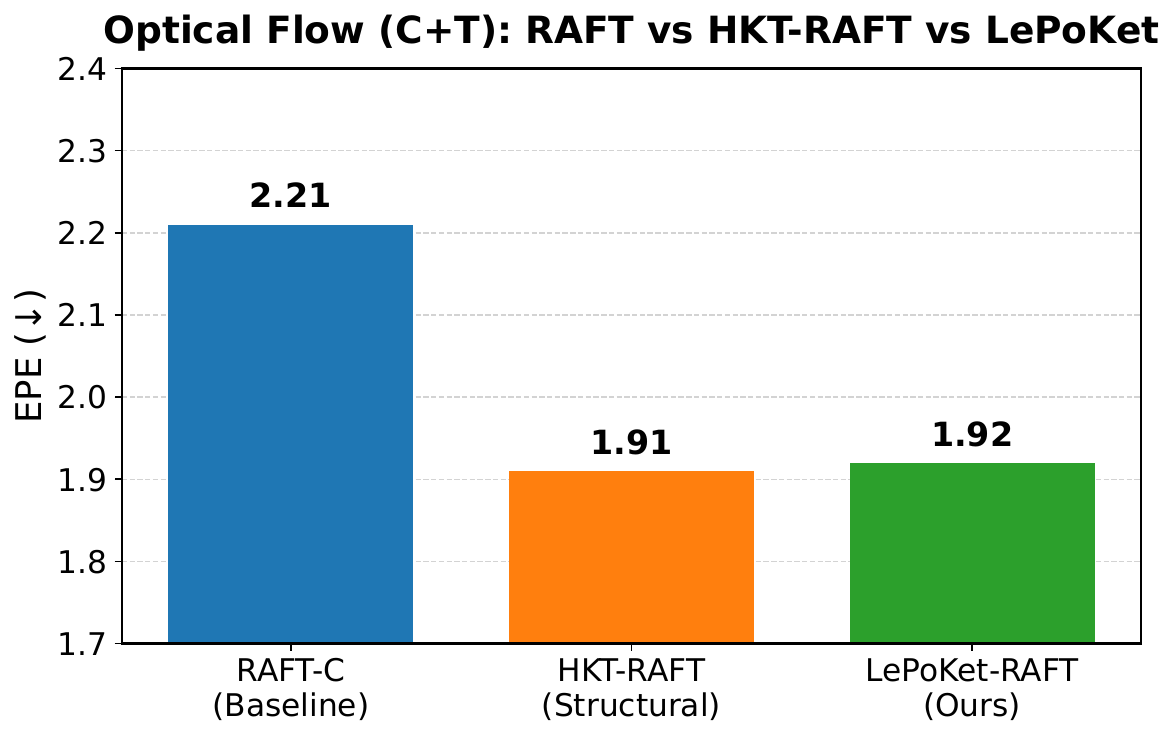}
        {\small (b) Optical flow (C+T)}
    \end{minipage}
   \caption{Structural vs.\ learnable knowledge transfer. (a) CIFAR-10 accuracy for ResNet110$\rightarrow$ResNet20. (b) Sintel Clean EPE under C+T training. Higher accuracy and lower EPE are better.}
    \label{fig:lepoket_comparison}
\end{figure}

On CIFAR-10, Fig.~\ref{fig:lepoket_comparison}(a) shows that HKT improves the baseline from 91.25\% to 92.40\%, corresponding to 13.14\% RER. LePoKet reaches 93.40\% and 24.57\% RER. Relative to HKT, learning the interaction yields an additional 1.00 percentage-point improvement in this experiment, showing a substantial benefit beyond the structural inheritance path alone.

The optical-flow ablation in Fig.~\ref{fig:lepoket_comparison}(b) is deliberately reported without overstating the effect of learnability. The published HKT results provide two C+T references: 2HKT-RAFT obtains 1.91/3.03/7.37 and 3HKT-RAFT obtains 1.90/3.08/6.45 on S-C/S-F/K-E. LePoKet obtains 1.92/3.01/6.39. Hence HKT retains a marginal advantage on Sintel Clean, whereas LePoKet is better on Sintel Final and KITTI EPE. Table~\ref{tab:flow_ct} also establishes that all structural-transfer variants improve the compact baseline on the principal reported EPE metrics. The evidence therefore supports structural transfer strongly, while the incremental value of learning the interaction is metric-dependent.

\subsection{Direct HKT--LePoKet Extension Analysis}
The HKT comparison is more than an additional baseline: it identifies what changes when structural inheritance becomes task-learnable. HKT established the ETM abstraction and reports 92.40\% on CIFAR-10 for HKT-ResNet20 while retaining the 0.27M-parameter child and approximately 0.10 ms inference time on its TITAN RTX setup \cite{tchenko2026hereditary}. Under the same ResNet110$\rightarrow$ResNet20 setting, LePoKet reaches 93.40\%. The 1.00-point improvement over HKT indicates that explicitly learning the transfer interface provides a substantial additional benefit in this classification setting.

Dense motion gives a complementary result. Under C+T training, 2HKT-RAFT reports 1.91/3.03/7.37 and 3HKT-RAFT 1.90/3.08/6.45 on S-C/S-F/K-E \cite{tchenko2026hereditary}; LePoKet obtains 1.92/3.01/6.39. Thus, HKT remains marginally better on Sintel Clean, whereas LePoKet is better on Sintel Final and KITTI EPE. Learnability therefore changes the cross-benchmark trade-off rather than uniformly improving every metric.

LePoKet also changes the optimization mechanism: HKT aggregates task and inheritance supervision, whereas LePoKet learns its transfer parameters through $\mathcal L_{\mathrm{task}}$ without an explicit teacher--student discrepancy. The present study does not yet isolate these two changes factorially, nor does it report LePoKet hardware latency. Consequently, HKT's published efficiency numbers are not reused as LePoKet measurements. A future ablation should independently vary fixed/learnable mixing and task-only/inheritance-augmented supervision, together with embedded-hardware profiling.

\subsection{Implications for Robotic Vision}
Optical flow is a dense representation of scene motion and is relevant to perception pipelines that reason about dynamic environments. The RAFT experiment is therefore the robotics-facing component of this study: it evaluates whether LePoKet can strengthen a compact dense predictor without adding a new teacher-imitation objective or fine-tuning on the evaluation benchmarks. The result is encouraging for resource-constrained perception, but it should not be interpreted as an end-to-end robot-system evaluation.

The present paper does not report on-device latency, energy consumption, memory use, or closed-loop navigation/control performance. Nor does it claim that the transfer interface is cost-free during training. These measurements are important for establishing deployment efficiency and constitute the main limitation of the current evaluation. A natural next step is to train with LePoKet and then characterize the compact child on embedded robotic hardware, as well as to evaluate transfer in downstream tasks such as navigation or scene understanding.

\subsection{Summary of Findings}
Across the reported experiments, Q1 is supported by improvements over both compact baselines; Q2 is supported by the transfer from ResNet classification to RAFT optical flow; and Q3 shows that structural inheritance provides a strong gain and that learning the transfer interface adds a further 1.00 percentage point on CIFAR-10, while the benefit remains metric-dependent in optical flow. This distinction is central to the interpretation of LePoKet: its
contribution is not an assertion that learned interaction must win on
every benchmark value, but a general mechanism that makes the
parameters governing structural knowledge transfer explicitly
optimizable by the downstream task.

\section{Conclusion}

We presented LePoKet, a learnable extension of the structural HKT paradigm \cite{tchenko2026hereditary} and a framework for transferring knowledge from a frozen parent network to a compact child model through the forward computation. Rather than introducing an auxiliary teacher--student imitation objective, LePoKet parameterizes an Extract--Transform--Mix interface and uses Learnable Genetic Attention to regulate residual feature inheritance jointly with child optimization. Controlled ResNet experiments on CIFAR-10 and CIFAR-100 show improvements over standard child training, while the RAFT-based study demonstrates that the same mechanism extends to dense motion estimation. Under C+T training without Sintel or KITTI fine-tuning, LePoKet improves the compact RAFT baseline on all three reported optical-flow metrics. The direct HKT comparison shows a further 1.00 percentage-point CIFAR-10 gain from learning the transfer interface (92.40\% for HKT versus 93.40\% for LePoKet). In optical flow, LePoKet slightly trails 2HKT/3HKT on Sintel Clean but improves Sintel Final and KITTI EPE, underscoring that learnability changes the cross-benchmark trade-off rather than necessarily improving every individual metric. Overall, the results motivate learnable structural transfer as a promising direction for efficient robotic perception. Future work should evaluate the framework directly on embodied robotic tasks and characterize its latency, memory, and computational overhead on deployment hardware.


\bibliographystyle{IEEEbib}
\bibliography{strings,refs}




\end{document}